\def\year{2021}\relax
\documentclass[letterpaper]{article} % DO NOT CHANGE THIS
\usepackage{aaai21}  % DO NOT CHANGE THIS
\usepackage{times}  % DO NOT CHANGE THIS
\usepackage{helvet} % DO NOT CHANGE THIS
\usepackage{courier}  % DO NOT CHANGE THIS
\usepackage[hyphens]{url}  % DO NOT CHANGE THIS
\usepackage{graphicx} % DO NOT CHANGE THIS
\usepackage{subfigure}
\usepackage{algorithm}
\usepackage{algorithmic}
\usepackage{multirow}
\usepackage{balance}
\usepackage{graphicx}  % DO NOT CHANGE THIS
\usepackage{natbib}  % DO NOT CHANGE THIS AND DO NOT ADD ANY OPTIONS TO IT
\usepackage{caption} % DO NOT CHANGE THIS AND DO NOT ADD ANY OPTIONS TO IT
\title{Deep Active Learning by Model Interpretability}

 \author{
     Qiang Liu\textsuperscript{\rm 1,2}, Zhaocheng Liu\textsuperscript{\rm 1}, Xiaofang Zhu\textsuperscript{\rm 1}, Yeliang Xiu\textsuperscript{\rm 3}\\
 }
 \affiliations{
     \textsuperscript{\rm 1}RealAI \textsuperscript{\rm 2}Tsinghua University \textsuperscript{\rm 3}Sun Yat-sen University\\
     \{qiang.liu,zhaocheng.liu,xiaofang.zhu\}@realai.ai,  1448560127@qq.com
 }
\begin{document}

\maketitle

\begin{abstract}
Recent successes of Deep Neural Networks (DNNs) in a variety of research tasks, however, heavily rely on the large amounts of labeled samples.
This may require considerable annotation cost in real-world applications.
Fortunately, active learning is a promising methodology to train high-performing model with minimal annotation cost.
In the deep learning context, the critical question of active learning is how to precisely identify the informativeness of samples for DNN.
In this paper, inspired by piece-wise linear interpretability in DNN, we introduce the linearly separable regions of samples to the problem of active learning, and propose a novel Deep Active learning approach by Model Interpretability (DAMI).
To keep the maximal representativeness of the entire unlabeled data, DAMI tries to select and label samples on different linearly separable regions introduced by the piece-wise linear interpretability in DNN.
We focus on modeling Multi-Layer Perception (MLP) for modeling tabular data.
Specifically, we use the local piece-wise interpretation in MLP as the representation of each sample, and directly run K-Center clustering to select and label samples.
To be noted, this whole process of DAMI does not require any hyper-parameters to tune manually.
To verify the effectiveness of our approach, extensive experiments have been conducted on several tabular datasets.
The experimental results demonstrate that DAMI constantly outperforms several state-of-the-art compared approaches.
\end{abstract}

\section{Introduction}

Over the past decades, Deep Neural Networks (DNNs) have represented the state-of-the-art supervised learning models and shown unprecedented success in numerous research tasks.
However, these successes heavily rely on large amount of labeled training samples.
A promising approach to address this problem is active learning, which aims to find effective ways to identify and label the maximally informative samples from a pool of unlabeled data \cite{wang2015querying,ash2019deep}.

Previous works on active learning mainly quantify samples from uncertainty and representative.
Expected Gradient Length (EGL) \cite{huang2016active,zhang2017active} is a typical uncertainty-based method, which regards the norms of gradients of losses with respect to the model parameters as the uncertainty evaluation.
Bayesian Active Learning by Disagreement (BALD) \cite{houlsby2011bayesian,gal2017deep,siddhant2018deep} measures uncertainty according to the probabilistic distribution of model outputs via Bayesian inference \cite{zhu2017big}, where an approximation by dropout are usually incorporated \cite{gal2016dropout}.
Among representative-based approaches, in the deep learning context, some works define the active learning task as a CORESET problem \cite{sener2017active}, which uses the representations of the last layer in DNN as representations of samples.
Besides, there are several approaches trade off between uncertainty and representative \cite{wang2015querying,ash2019deep}.
For the active learning task in deep learning, Batch Active learning by Diverse Gradient Embeddings (BADGE) \cite{ash2019deep} utilizes gradients of losses with respect to the representations of the last layer in DNN as representations of samples, on which clustering is conducted for capturing both uncertainty and representative.

Recently, the interpretability of DNN has been widely studied, among which most works focus on local piece-wise interpretability \cite{ribeiro2016should,chu2018exact}.
Specifically, the local piece-wise interpretations of DNN can be calculated via gradient backpropagation \cite{li2015visualizing,selvaraju2017grad} or feature perturbation \cite{fong2017interpretable,guan2019towards}.
Some previous works \cite{montufar2014number,harvey2017nearly,chu2018exact} deeply investigate the local interpretability of DNN, and show that DNN with piece-wise linear activations, e.g., Maxout \cite{goodfellow2013maxout} and the family of ReLU \cite{nair2010rectified,glorot2011deep}, can be regraded as a set of numerous local linear classifiers.
That is to say, with DNN, samples are divided into numerous linearly separable regions, and all samples in the same linearly separable region are classified by the same local linear classifier \cite{chu2018exact}.
As we know, we usually need the same numbers of samples for fitting different linear classifiers in different linearly separable regions.
Thus, to select samples for optimally training DNN, different linearly separable regions should be considered in a balance way.
From this perspective, with the help of local interpretability of DNN, we can identify different linearly separable regions of samples, and potentially promote the effectiveness of deep active learning.

Accordingly, in this paper, we introduce the linearly separable regions of samples to the problem of active learning for DNN, and propose a novel Deep Active learning approach by Model Interpretability (DAMI).
Specifically, we calculate the local interpretations in DNN via the gradient backpropagation from the final predictions to the input features \cite{li2015visualizing,selvaraju2017grad}.
In this paper, we focus on Multi-Layer Perception (MLP) for classification on tabular data.
Specifically, we use local interpretations in MLP as the representations of samples, and directly run K-Center \cite{sener2017active} clustering to select and label samples.
We have conducted extensive experiments on four tabular datasets.
The experimental results show that DAMI can constantly outperform state-of-the-art active learning approaches.

\section{Related Works}

In this section, we briefly review some related works on active learning, as well as interpretability of DNN.

\subsection{Active Learning}

Based on a certain sampling strategy, active learning approaches actively samples a small batch of informative instances from the unlabeled data for labeling.
Roughly speaking, there exist two major types of strategies: representative-based sampling and uncertainty-based sampling.

Representative-based sampling aims to select unlabeled samples that are representative according to the data distribution.
In the deep learning context, this is usually done based on CORESET construction \cite{sener2017active}, in which the representations of the last layer in DNN are used as representations of samples.
Adversarial learning can also be considered to select most indistinguishable samples \cite{ducoffe2018adversarial}.

Uncertainty-based sampling aims to select samples that can maximally reduce the uncertainty of the classifier.
Such approaches are widely applied in the deep learning context.
EGL \cite{huang2016active} measures uncertainty based on the norms of gradients of losses with respect to the model parameters.
For the task of sentence classification, EGL-word \cite{zhang2017active} seeks to find the word with largest norm of gradients in a sentence, and uses the corresponding norm as the uncertainty measurement.
BALD \cite{houlsby2011bayesian} measures uncertainty according to the probabilistic distribution of model outputs via Bayesian inference \cite{zhu2017big}.
Inspired by the finding that Bayesian inference can be approximated by dropout in deep models \cite{gal2016dropout}, the dropout approximation is usually applied in deep active learning to perform BALD \cite{gal2017deep}.
And the BALD approach is successfully applied in the task of sentence classification \cite{siddhant2018deep}.
Meanwhile, the uncertainty-based approaches have been empirically studied and evaluated for deep active learning on textual data \cite{prabhu2019sampling}.

Furthermore, some works consider uncertainty and representative at the same time, and make trade-off between them \cite{huang2010active,wang2015querying,hsu2015active}.
For example, such trade-off is considered for text classification \cite{yan2020active}.
In the context of deep learning, BADGE \cite{ash2019deep} is proposed to take use of gradients of losses with respect to the representations of the last layer in DNN as representations of samples, in which both uncertainty and representative can be preserved to some extend.
BADGE can be viewed as a combination of CORESET and EGL.

\begin{figure}
\centering
\includegraphics[width=0.4\textwidth]{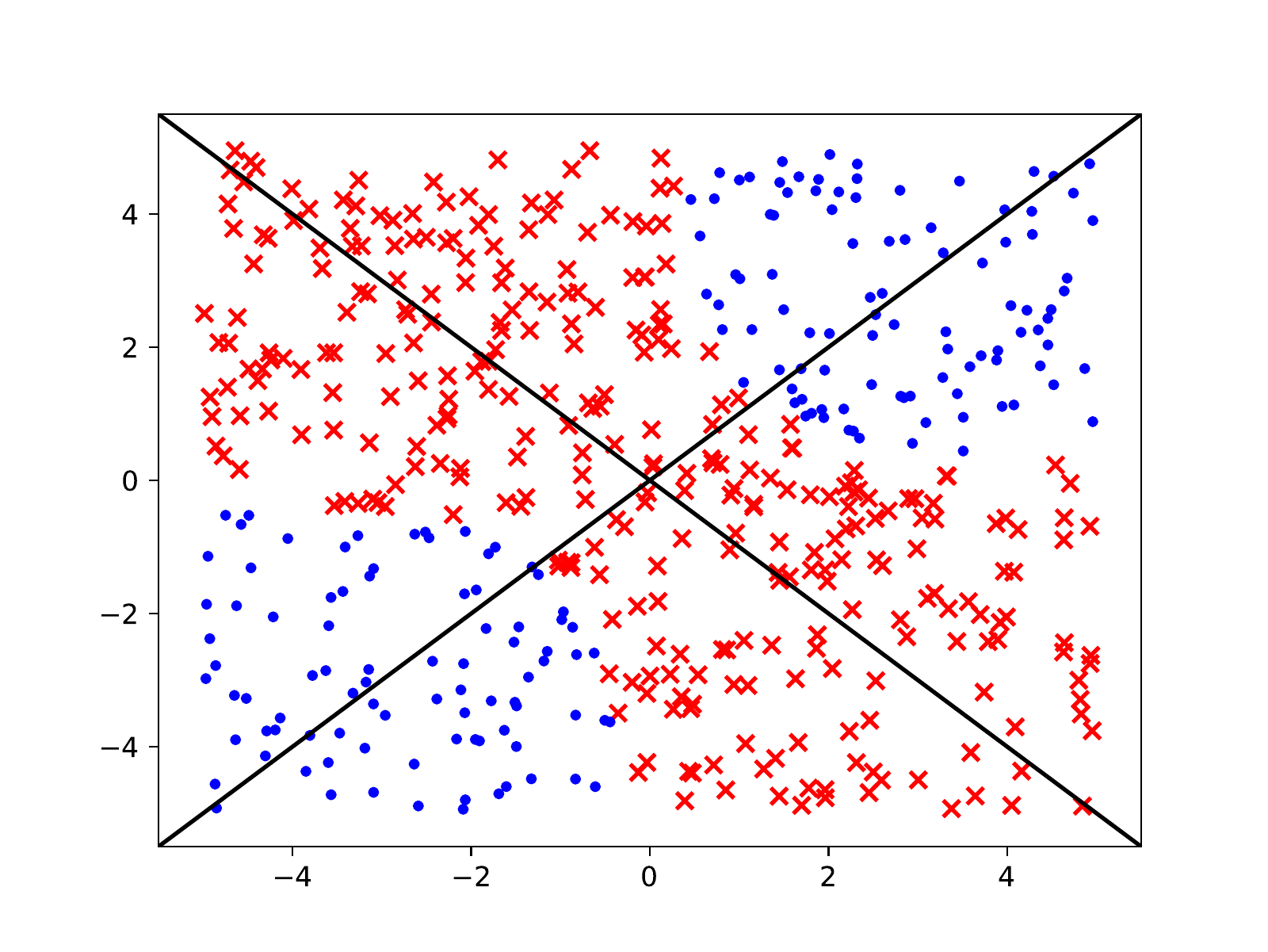}
\caption{Distribution of example data in the Sigmoid dataset, which is formulated in Eq. (\ref{equation:toy_data}). There are mainly four linearly separable regions, which are shown in the triangles.}
\label{fig:distribution}
\end{figure}

\begin{figure*}
	\centering
	\subfigure[CORESET.]{
		\begin{minipage}[b]{0.32\textwidth}
			\includegraphics[width=1\textwidth]{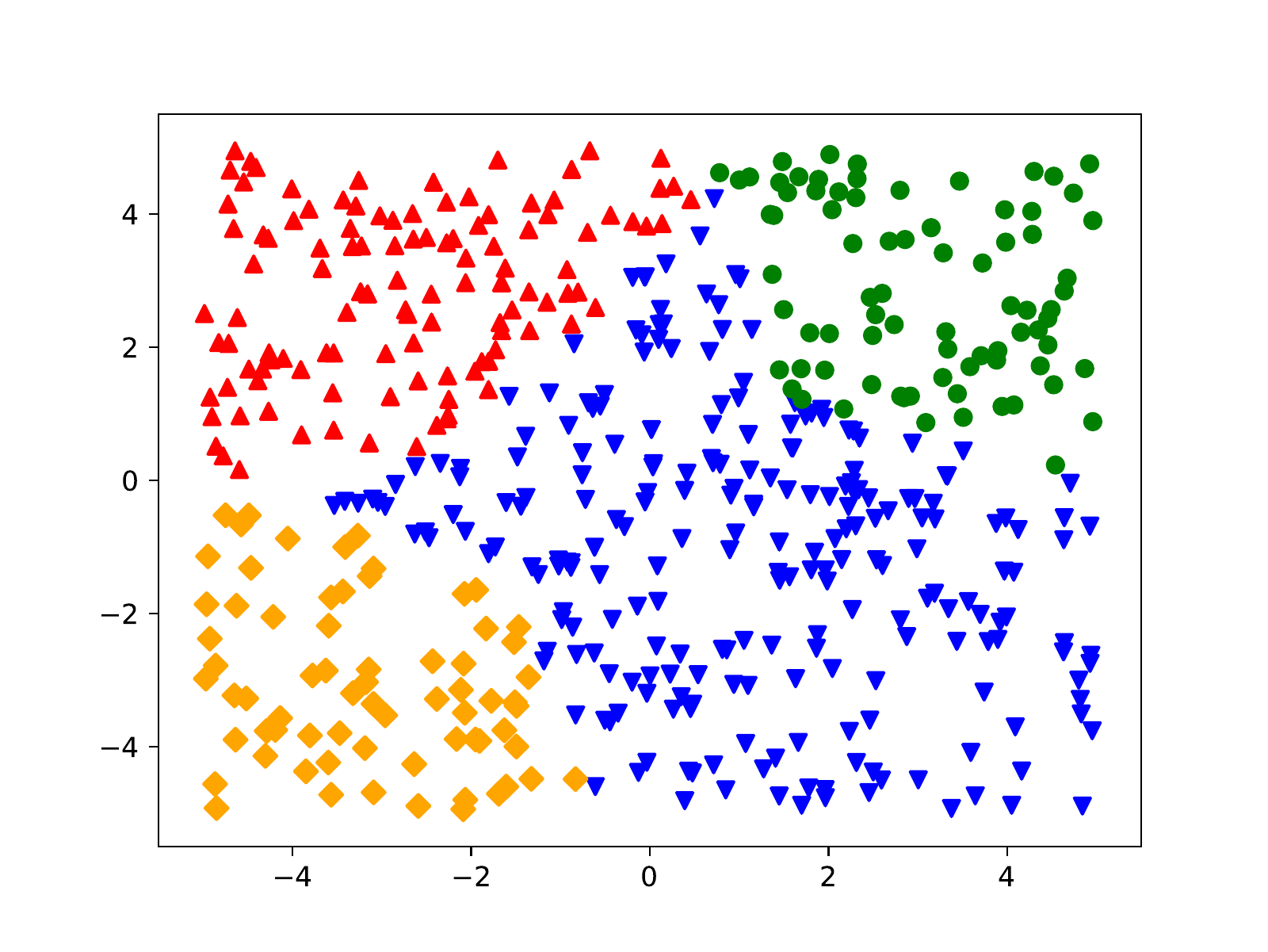}
		\end{minipage}
	}
	\subfigure[BADGE.]{
		\begin{minipage}[b]{0.32\textwidth}
			\includegraphics[width=1\textwidth]{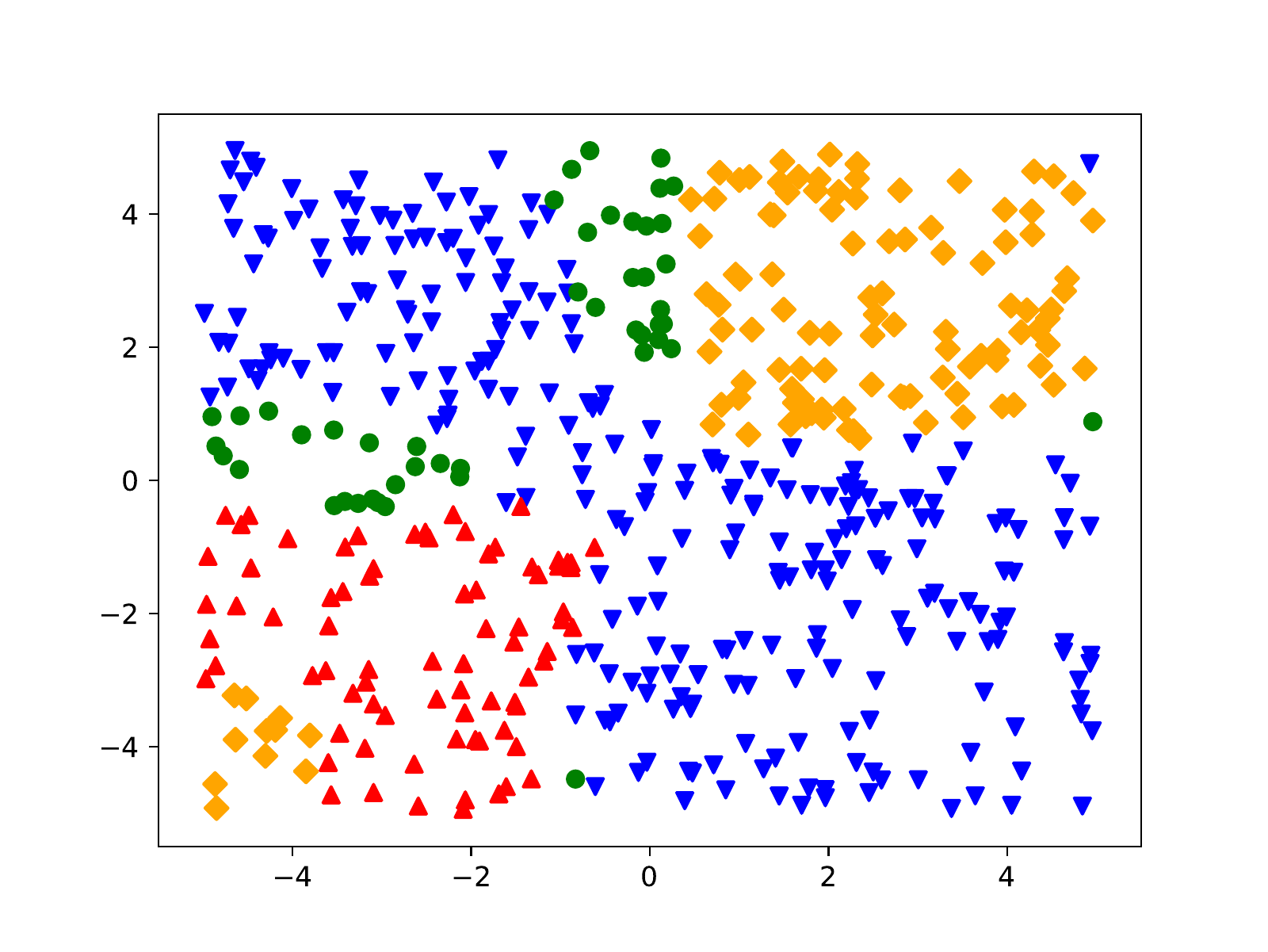}
		\end{minipage}
	}
	\subfigure[local interpretations.]{
		\begin{minipage}[b]{0.32\textwidth}
			\includegraphics[width=1\textwidth]{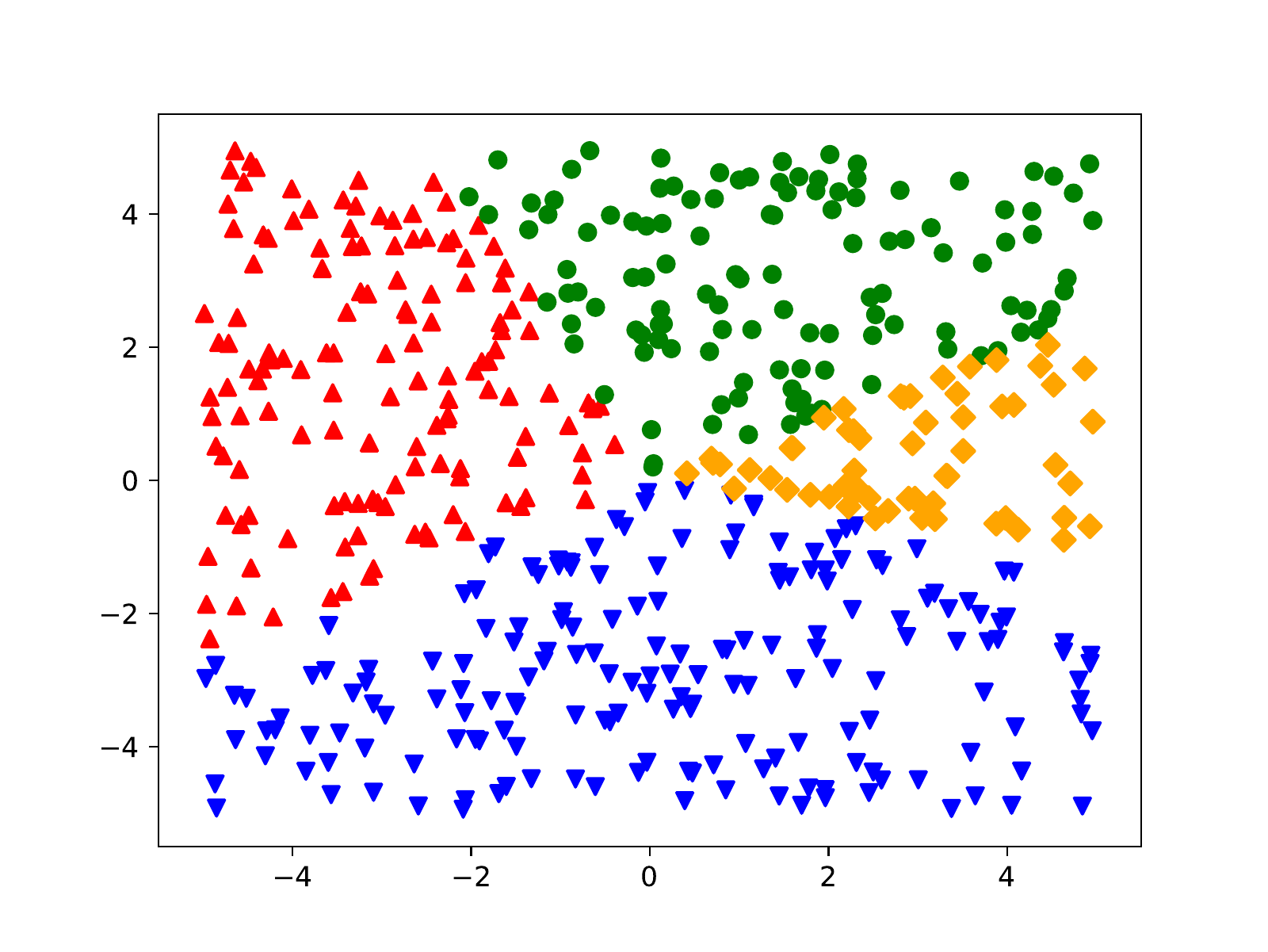}
		\end{minipage}
	}
	\caption{K-means clustering on samples in the Sigmoid dataset, which is formulated in Eq. (\ref{equation:toy_data}) and illustrated in Fig. \ref{fig:distribution}. The representations of samples are from CORESET \cite{sener2017active} and BADGE \cite{ash2019deep}, as well as the local interpretations in DNN via the calculation in Eq. (\ref{equation:local_interpretation}). Clearly, only the local interpretations can find the four linearly separable regions in the Sigmoid dataset.}
	\label{fig:cluster}
\end{figure*}

\subsection{Interpretability of DNN}

Recently, the interpretability of DNN has drawn great attention in academia, and research works mostly focus on local piece-wise interpretability, which means assigning a piece of local interpretation for each sample \cite{guidotti2018survey}.
Some unified approaches are proposed to fit a linear classifier in each local space of input samples \cite{ribeiro2016should,lundberg2017unified}.
Some works investigate the gradients from the final predictions to the input features in deep models, which can be applied in the visualization of deep vision models \cite{zhou2016learning,selvaraju2017grad,smilkov2017smoothgrad,alvarez2018towards}, as well as the interpretation of language models \cite{li2015visualizing,yuan2019interpreting}.
Perturbation on input features is also utilized to find local interpretations of both vision models \cite{fong2017interpretable} and language models \cite{guan2019towards}.
Meanwhile, via adversarial diagnosis of neural networks, adversarial examples can also be introduced for local interpretation of DNN \cite{koh2017understanding,dong2018towards}.
In some views, attention in deep models can also be regarded as local interpretations  \cite{wang2019towards,sun2020understanding}.
% Furthermore, some works try to find feature interactions in DNN \cite{tsang2020feature}, or understand the structures in deep language models \cite{tenney2019you,hao2019visualizing,wu2020perturbed}.

As discussed in some previous works \cite{ribeiro2016should,lundberg2017unified}, the nonlinear DNN model can be regarded as a combination of numbers of linear classifiers.
And the upper bound of the number of linear classifiers in DNN with piece-wise linear activation functions, e.g., Maxout \cite{goodfellow2013maxout} and the family of ReLU \cite{nair2010rectified,glorot2011deep}, has been given \cite{montufar2014number}.
Moreover, piece-wise linear DNN has been exactly and consistently interpreted as a set of linear classifiers \cite{chu2018exact}.
In a word, with the local piece-wise interpretations of DNN, we can define the linearly separable regions of input samples.

\section{The DAMI Approach}

In this section, we introduce the proposed DAMI approach for the MLP model for the classification task on tabular data.

\subsection{Notations}

In this work, we consider the pool-based AL case \cite{tong2001support,settles2008analysis,zhang2017active}, in which we have a small set of labeled samples $\mathcal L$, and a large set of unlabeled samples $\mathcal U$.

For sample $s_i \in \mathcal L$, we have $s_i = \left( {x_i,y_i} \right)$, where $x_i$ and $y_i \in \{ 0,1 \}$ are the corresponding features and label.
For sample $s_i \in \mathcal U$, we have $s_i = \left( {x_i,} \right)$, where the label is unknown.
For sample $s_i$ in tabular data, $x_i$ is a fixed-size feature vector, and denoted as $x_i = \left( x_{i,1}, x_{i,2},...,x_{i,M} \right)$, where $M$ is the width of the tabular data.
With the labeled samples $\mathcal L$, we can train a deep learning-based classifier $f\left( x|\theta \right)$: $\mathcal X \to \mathcal Y$, which maps the features to the labels.
Then, we can develop a AL strategy to select most informative samples from $\mathcal U$, and further optimize the classifier.

\subsection{Learning from the Interpretations of DNN}

Recently, extensive works have been conducted to study local piece-wise interpretability of DNN.
The calculation of local interpretations can be done via the gradient backpropagation from the predictions to the input features \cite{selvaraju2017grad,li2015visualizing,smilkov2017smoothgrad,yuan2019interpreting}.
We need to first train a deep model, and obtain the predicted label ${\hat y}_i$ for sample $s_i \in \mathcal U$ or $s_i \in \mathcal L$.
Then, we can calculate local interpretations of sample $s_i$ as
\begin{equation} \label{equation:local_interpretation}
\mathop I\nolimits_{i}  = \frac{{\partial \mathop {\hat y}\nolimits_i }}{{\partial \mathop x\nolimits_{i} }}.
\end{equation}
As in \cite{li2015visualizing}, we can have the following form of local interpretation
\begin{equation}
\mathop {\hat y}\nolimits_i  \approx  {\mathop I\nolimits_i \mathop x\nolimits_i^\top  + b}.
\end{equation}
As mentioned in some works \cite{montufar2014number,ribeiro2016should,chu2018exact}, a DNN model with piece-wise linear activation functions \cite{goodfellow2013maxout,nair2010rectified,glorot2011deep} can be regarded as a combination of numbers of linear classifiers, where local interpretations $I_i$ are the weights of the linear classifiers.
That is to say, with the local piece-wise interpretations in DNN, samples can be divided into numerous linearly separable regions, and samples in the same linearly separable region are classified by the same local linear classifier \cite{chu2018exact}.
Thus, local interpretations of samples as calculated in Eq. (\ref{equation:local_interpretation}) can be partitioned into several clusters, and each of them corresponds to the linear classifiers in a specific linearly separable region.
We usually need the same numbers of samples for fitting different linear classifiers in different linearly separable regions.
Accordingly, to select samples for optimally training DNN in deep active learning, different linearly separable regions should be considered in a balance way.

\begin{figure*}
\centering
\includegraphics[width=0.93\textwidth]{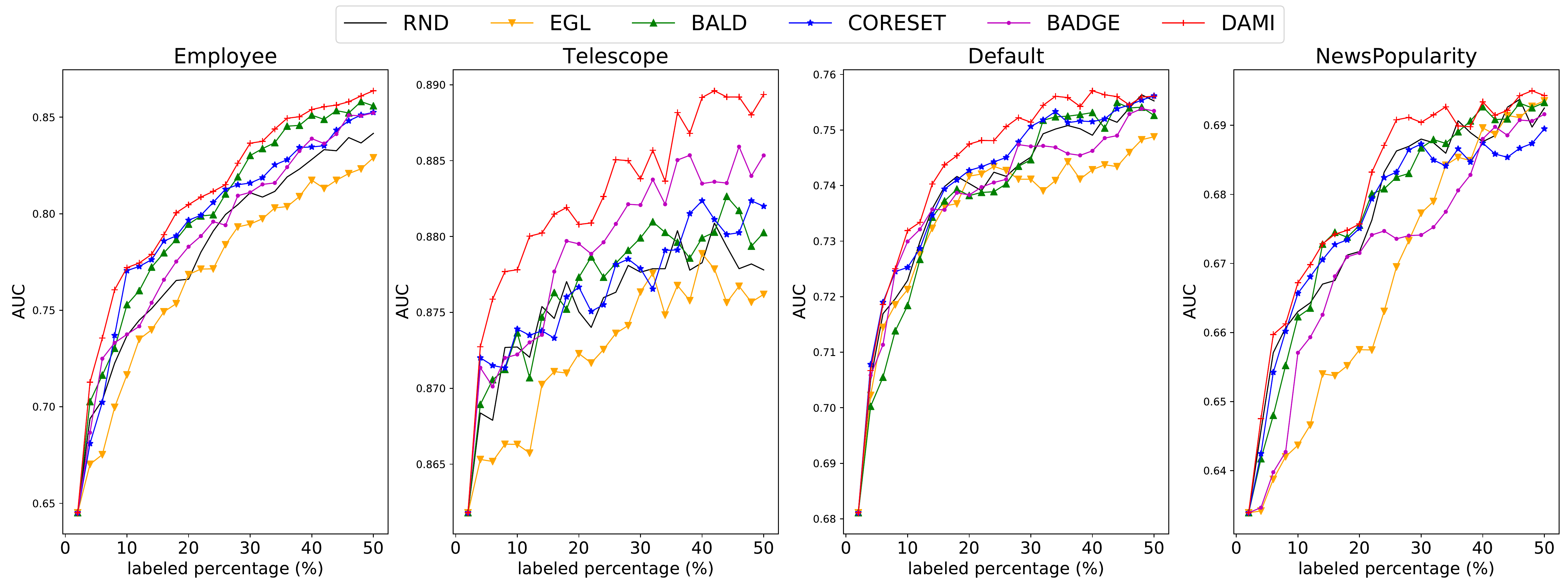}
\caption{Performance comparison with different ratios of labeled samples on tabular datasets.}
\label{fig:curve_tabular}
\end{figure*}

To demonstrate the local interpretations of DNN can help to promote deep active learning, we draw some example data from the following probability distribution
\begin{equation} \label{equation:toy_data}
p\left( {\mathop y\nolimits_i  = 1|\mathop x\nolimits_i } \right) = \sigma \left( {\mathop x\nolimits_{i,1} * \mathop x\nolimits_{i,2} } \right),
\end{equation}
where $x_{i,1}$ and $x_{i,2}$ are uniformly sampled from $[ -5.0,5.0 ]$, and $\sigma \left( \cdot \right)$ is the sigmoid function.
For simplicity, these toy samples are named the Sigmoid dataset, whose data distribution is shown in Fig. \ref{fig:distribution}.
This data is clearly nonlinear, and there are roughly four linearly separable regions, which are illustrated in the four triangles.
With the help of the interpretations of DNN, we are able to find different linearly separable regions of the unlabeled samples, and propose a better deep active learning approach.
To illustrate this, for samples in the Sigmoid dataset, we run K-means clustering on the representations generated by CORESET \cite{sener2017active} and BADGE \cite{ash2019deep}, as well as the local interpretations in a MLP model trained on the Sigmoid dataset.
We set the number of clusters in K-means as $4$, and results are shown in Fig. \ref{fig:cluster}.
We can observe that, CORESET focuses on the original feature distribution and different classes, while BADGE pays more attention to the decision boundaries.
And clearly, we can only use local interpretations to find the four linearly separable regions.

\begin{algorithm}[t]
    \caption{DAMI on Tabular Data.}
    \label{alg:DAMI1}
    \small
    \begin{algorithmic}[1]
        \REQUIRE Labeled samples $\mathcal L$, unlabeled samples $\mathcal U$, number of iterations $M$, budget $k$ in each iteration of sample selection.
        \STATE Train an initial MLP model $f\left( x|\theta_0 \right)$ on $\mathcal L$;
        \FOR{$m=1,2,...,M$}
            \FOR{$s_i \in \mathcal U$}
                \STATE Make prediction $\mathop {\hat y}\nolimits_i  = f\left( {\mathop x\nolimits_i |\mathop \theta \nolimits_{m - 1} } \right)$;
                \STATE Compute the local interpretation $I_i$ as in Eq. (\ref{equation:local_interpretation});
            \ENDFOR
            \STATE Run K-Center on $\left\{ {\mathop I\nolimits_i |\mathop s\nolimits_i  \in \mathcal L} \right\}$ and $\left\{ {\mathop I\nolimits_i |\mathop s\nolimits_i  \in \mathcal U} \right\}$ to find $k$ samples in $\mathcal U$ for labeling as $\mathcal L_m$;
            \STATE Label samples $s_i \in \mathcal L_m$;
            \STATE $\mathcal L \leftarrow \mathcal L \cup \mathcal L_m$;
            \STATE $\mathcal U \leftarrow \mathcal U \backslash \mathcal L_m$;
            \STATE Train a new MLP model $f\left( x|\theta_m \right)$ on $\mathcal L$;
        \ENDFOR
        \RETURN The final model $f\left( x|\theta_M \right)$.
    \end{algorithmic}
\end{algorithm}

\subsection{DAMI on Tabular Data}

Inspired by the local interpretability of DNN, we can introduce piece-wise linearly separable regions to the problem of deep active learning.
Specifically, in this subsection, we detail the DAMI approach on tabular data.

In tabular data, there are fixed number of input features, and MLP is usually applied for modeling.
Thus, we first train a MLP model on tabular data.
Then, to find most informative unlabeled samples, according to the calculation in Eq. (\ref{equation:local_interpretation}), we can directly utilize the local interpretations as the representations of samples.
K-Center has been successfully used for finding informative samples based on their representations \cite{sener2017active}.
Accordingly, we also adopt K-Center in our approach.
Specially, with budget $k$ in each iteration of sample selection, we run K-Center clustering on $\left\{ {\mathop I\nolimits_i |\mathop s\nolimits_i  \in \mathcal L} \right\}$ and $\left\{ {\mathop I\nolimits_i |\mathop s\nolimits_i  \in \mathcal U} \right\}$ to find $k$ samples in $\mathcal U$ for labeling.
Detailed process can be found in Alg. \ref{alg:DAMI1}.

\begin{table}[t]
  \centering
  \small
  \caption{Details about tabular datasets.}
    \begin{tabular}{c|cccc}
    \hline
    dataset & \#samples & \#positive & \#negative & \#features \\
    \hline
    Employee & 32769 & 30872 & 1897  & 9 \\
    Telescope & 19020 & 6688  & 12332 & 11 \\
    Default & 30000 & 6636  & 23364 & 24 \\
    NewsPopularity & 39644 & 2215  & 37429 & 61 \\
    \hline
    \end{tabular}%
  \label{tab:tabular}%
\end{table}%

\section{Experiments}
In this section, we empirically evaluate our proposed DAMI approach on tabular data.

\subsection{Experiments on Tabular Datasets}

To evaluate the performances of DAMI on tabular data, we conduct comparison among following approaches:
\begin{itemize}
\item \textbf{RND} is a simple baseline which randomly selects samples in each iteration.
\item \textbf{EGL} \cite{huang2016active} is a typical uncertainty-based approach, which utilizes norms of gradients.
\item \textbf{BALD} \cite{houlsby2011bayesian} is a another uncertainty-based approach based on Bayesian inference. We apply dropout approximation \cite{gal2016dropout,gal2017deep} in our experiments.
\item \textbf{CORESET} \cite{sener2017active} uses the representations of the last layer in DNN as the representations.
\item \textbf{BADGE} \cite{ash2019deep} can be viewed as a combination of EGL and CORESET.
\item \textbf{DAMI} is proposed in this paper, which conduct deep active learning based the local interpretability in DNN.
\end{itemize}
We run $3$ layers of MLP with ReLU activation on samples in each dataset, where the hidden units are set as $\left(16, 8\right)$ and the dropout rate is set as $0.8$.
We involve four tabular datasets: \textbf{Employee}, \textbf{Telescope}, \textbf{Default} and \textbf{NewsPopularity}.
Details about these datasets can be found in Tab. \ref{tab:tabular}.
Considering these datasets are class-imbalanced, we use AUC (Area Under Curve) as the evaluation metric.
We randomly select $60\%$, $20\%$ and $20\%$ samples in each dataset for training, validation and testing respectively.
We use $2\%$ samples in the training set as initial labeled samples.
Then, we label $2\%$ samples in the training set during each iteration of sample selection, until $50\%$ samples in the training set are covered.
We run each approach $10$ times, and report the median of  experimental results.

Fig. \ref{fig:curve_tabular} shows the performance comparison among RND, EGL, BALD, CORESET, BADGE and DAMI with different ratios of labeled samples.
In most cases, active learning approaches can outperform the random selection, which demonstrates the necessity of deep active learning.
We can observe that, EGL performs poor, and is even outperformed by RND.
This may indicate that, the uncertainty evaluation based the norms of gradients is not stable.
On the Employee, Telescope and Default datasets, BALD, CORESET and BADGE have close performances, and each of them achieves the best performance among the five baseline methods on different datasets.
Meanwhile, BADGE performs poor on the NewsPopularity dataset.
Moreover, it is clear that, DAMI has best performances on the four tabular datasets, and can constantly outperform other baseline approaches.
Specifically, in the middle parts of the curves, i.e., labeled percentage in the range of $[15\%,35\%]$, DAMI usually has great advantages.

\section{Conclusion}

In this paper, inspired by the local piece-wise interpretability of DNN, we introduce the linearly separable regions of samples to the problem of active learning.
Accordingly, we propose a novel DAMI approach, which selects and labels samples on different linearly separable regions for optimally training DNN.
We mainly focus the scenarios of MLP for classification on tabular data.
Specifically, we use the local piece-wise interpretation in DNN as the representation of each sample, and directly run K-Center clustering to select and label samples.
Extensive experiments on four tabular demonstrate the effectiveness of our proposed DAMI approach.

\balance
\bibliography{DAMI}

\end{document}